\documentclass{article}

\usepackage[english]{babel}

\usepackage[letterpaper,top=2cm,bottom=2cm,left=3cm,right=3cm,marginparwidth=1.75cm]{geometry}

\usepackage[colorlinks=true, allcolors=blue]{hyperref}

\usepackage{geometry}
\usepackage{amsmath}
\usepackage{amsthm}
\usepackage{amssymb}
\usepackage{graphicx}
\usepackage{hyperref}
\usepackage{xcolor}
\usepackage{esvect}
\usepackage[shortlabels]{enumitem}
\usepackage{ctable}
\usepackage{tabularx}
\usepackage{multirow}
\usepackage{booktabs}
\usepackage{subcaption}
\usepackage{mlmath}

\newtheorem*{thm*}{Theorem}
\newtheorem{thm}{Theorem}

\newtheorem{defi}{Definition}

\newtheorem{innercustomgeneric}{\customgenericname}
\providecommand{\customgenericname}{}
\newcommand{\newcustomtheorem}[2]{%
  \newenvironment{#1}[1]
  {%
   \renewcommand\customgenericname{#2}%
   \renewcommand\theinnercustomgeneric{##1}%
   \innercustomgeneric
  }
  {\endinnercustomgeneric}
}

\newcustomtheorem{customthm}{Theorem}
\newcustomtheorem{customlemma}{Lemma}

\usepackage{authblk}

\makeatletter
    \def\blfootnote{\xdef\@thefnmark{}\@footnotetext}
\makeatother

\title{An overview of condensation phenomenon in deep learning}

\author[1,2,*]{Zhi-Qin John Xu}
\author[1,2]{Yaoyu Zhang}
\author[1]{Zhangchen Zhou}

\affil[1]{School of Mathematical Sciences, Institute of Natural Sciences, MOE-LSC, Shanghai Jiao Tong University}
\affil[2]{School of Artificial Intelligence, Shanghai Jiao Tong University}
\affil[*]{Corresponding author: xuzhiqin@sjtu.edu.cn. Authors are listed in alphabetical order of last names.}

\begin{document}
\maketitle


\begin{abstract}
In this paper, we provide an overview of a common phenomenon, condensation, observed during the nonlinear training of neural networks: During the nonlinear training of neural networks, neurons in the same layer tend to condense into groups with similar outputs. Empirical observations suggest that the number of condensed clusters of neurons in the same layer typically increases monotonically as training progresses. Neural networks with small weight initializations or Dropout optimization can facilitate this condensation process. We also examine the underlying mechanisms of condensation from the perspectives of training dynamics and the structure of the loss landscape. 
The condensation phenomenon offers valuable insights into the generalization abilities of neural networks and correlates to stronger reasoning abilities in transformer-based language models.
\end{abstract}

\section{Introduction}

Deep neural networks (DNNs) have demonstrated remarkable performance across a wide range of applications. In particular, scaling laws suggest that improvements in performance for Large Language Models (LLMs) are closely tied to the size of both the model and the dataset \cite{kaplan2020scaling}. Understanding how these large-scale neural networks achieve such extraordinary performance is crucial for developing principles that guide the design of more efficient, robust, and computationally cost-effective machine learning models.

However, the study of large neural networks presents significant challenges, such as their enormous parameters and complex network architectures. Additionally, the data—ranging from language to image data—are often too complex to analyze using traditional methods. In this context, a phenomenon-driven approach has proven to be effective in uncovering insights into the behavior of neural networks.

One such phenomenon is the over-parameterization puzzle, which has led to a deeper understanding of neural network generalization \cite{breiman1995reflections,zhang2016understanding}. This puzzle reveals that a neural network can generalize well even when the number of parameters far exceeds the number of training data points. This observation challenges traditional learning theory, which typically improves generalization by imposing constraints on model complexity \cite{vapnik2013nature}. In contrast, the generalization of large neural networks appears to be largely independent of superficial complexity, such as the number of parameters. Instead, the optimization trajectory plays a crucial role in locating a minimum with specific properties among various types of minima. Empirical studies have shown that smaller batch sizes in Stochastic Gradient Descent (SGD) tend to lead to flatter minima, which is associated with better generalization \cite{keskar2016large}. This led to the development of sharpness-aware minimization (SAM) \cite{foretsharpness} techniques that further improve generalization performance. Additionally, recent works have shown that the noise covariance induced by SGD aligns with the Hessian of the loss landscape \cite{zhu2018anisotropic,wu2018sgd,feng2021inverse}, providing further insights into the optimization dynamics.

Another important empirical finding is the existence of a simplicity bias during neural network training \cite{arpit2017closer}. A series of experiments, followed by theoretical analysis, has identified a low-frequency bias, known as the frequency principle \cite{xu2019frequency,xu2024overview} or spectral bias \cite{rahaman2018spectral}, which helps explain the observed differences in generalization performance. This principle has also inspired the development of multi-scale DNN architectures \cite{liu2020multi,li2020multi,cai2019phasednn} and Fourier feature networks \cite{tancik2020fourier}, which accelerate the learning of high-frequency components in the data.

To further investigate the simplicity bias, several studies have analyzed the evolution of neural network parameters during training. Two distinct regimes \cite{luo2021phase,zhou2022empirical} have been identified: the linear regime, in which parameters initialized with relatively large values undergo minimal changes during training, and the nonlinear regime, where smaller initializations result in more substantial parameter adjustments \cite{rotskoff2018parameters,chizat2018global}. In the linear regime, the behavior of the neural network closely resembles that of kernel methods, with the neural tangent kernel (NTK) \cite{jacot2018neural,chizat2019lazy} being a prominent example. The transition between the linear and nonlinear regimes represents a critical phase, with mean-field dynamics being a typical example \cite{mei2019mean,sirignano2020mean,rotskoff2018parameters}. It is in the nonlinear regime that a universal condensation phenomenon occurs \cite{luo2021phase,zhou2022towards,zhou2022empirical}. In this paper, we aim to overview previous works on this phenomenon and provide a unified description of condensation:

\textbf{Condensation}: During the nonlinear training of neural networks, neurons in the same layer tend to condense into groups with similar outputs.

This condensed regime represents a state in which neurons in the same layer condense into a few distinct groups, with neurons within each group performing similar functions. This clustering phenomenon implies that a wide neural network can behave similarly to a much narrower network. Early in the nonlinear training process, neurons in the same layer tend to group into a small number of clusters \cite{maennel2018gradient, phuong2021the,lyu2021gradient,boursier2022gradient,zhou2022towards,zhou2022empirical,min2024early,wang2024understanding}. As training progresses, the number of clusters increases, which facilitates fitting. Thus, the condensation phenomenon offers a mechanism for the increasing complexity of the network’s outputs as training progresses.

In this paper, we present experiments with various neural network architectures to demonstrate the ubiquity of the condensation phenomenon in nonlinear training \cite{luo2021upper,zhou2022towards,zhou2023understanding}. We also explore how dropout \cite{srivastava2014dropout} implicitly induces a bias toward condensation \cite{zhang2024implicit,zhang2024stochastic}. Furthermore, we examine the origins of condensation from the perspectives of loss landscapes and training dynamics. The condensation phenomenon suggests a potential pruning strategy, where network size can be reduced without sacrificing generalization ability \cite{zhang2021embedding,chen2024efficient}. This insight also leads to a novel optimistic estimation of the sample size required to recover a target function based on a perfectly condensed network \cite{zhang2023optimistic} rather than relying on superficial network complexity, where the latter often leads to overly conservative estimates.

Moreover, the condensation phenomenon, originally observed in simple two-layer neural networks, provides a deeper understanding of the reasoning and memorization processes in transformer models, particularly for language tasks \cite{zhang2024initialization,zhang2025complexity}. This understanding could also inform methods for training transformer networks with improved reasoning capabilities.

Given that condensation is a prominent feature of the nonlinear training of neural networks, a deep understanding of this phenomenon would significantly enhance our comprehension and more effective utilization of deep learning.

This phenomenon has been characterized in various ways throughout the literature. \cite{maennel2018gradient} described it as a quantization effect where weight vectors tend to concentrate in finite directions due to gradient descent. \cite{brutzkus2019larger} referred to it as the weight clustering effect. \cite{chizat2019lazy} provided an illustrative example of non-lazy training. \cite{phuong2021the} named this behavior a form of inductive bias. Several works investigated this behavior of neurons within the same layer and named it ``alignment/get align''\cite{ji2018gradient,lyu2021gradient,boursier2022gradient,chistikov2023learning,min2024early,boursier2024early}. \cite{kumar2024directional,kumar2024early} termed this phenomenon ``directional convergence''.

\section{Concept of condensation}
The concept of condensation refers to the tendency of neurons within the same layer to condense into groups with similar outputs during training. This alignment or clustering of neurons is influenced by various hyperparameters and optimization methods, which can modulate the degree to which this similarity occurs. The similarity between neurons can be quantified using different metrics. Below, we present two such examples.

For a two-layer neural network with one-dimensional input:
\begin{equation}\label{eq:two_layer_network}
    h(x) = \sum\limits_{k=1}^{m}a_k\sigma(w_k x+b_k),
\end{equation}
the feature of the neuron $k$ is defined as $(\theta_k,A_k)$, where $\theta_k = \operatorname{sign}(b_k) \times \arccos\left(\frac{w_k}{\sqrt{w_k^2+b_k^2}}\right)$ and $A_k = \sqrt{w_k^2+b_k^2}$. 

By visualizing the two-dimensional features of all neurons during the training, it is ready to observe the condensation of such a simple network in a non-linear training process.

The aforementioned method is not suitable for visualizing neurons with high-dimensional inputs, such as those in the first hidden layer, which receives high-dimensional input vectors, or neurons in deeper layers, which process the outputs of multiple neurons from preceding layers. To address this, we can define the cosine similarity between the high-dimensional weight vectors of two neurons as a measure of their similarity.

\textbf{Cosine similarity:} The cosine similarity between two vectors $\vu_1$ and $\vu_2$ is defined as
\begin{equation}
    D(\vu_1,\vu_2) = \frac{\vu_1^\T\vu_2}{(\vu_1^\T\vu_1)^{1/2}(\vu_2^{\T}\vu_2)^{1/2}}.
\end{equation}
Two vectors have the same (or opposite) directions when their cosine similarity $D(\vu_1,\vu_2)$ is $1$ (or $-1$). 

For the activation function $\ReLU (x)=\max ({0,x})$, two neurons, with cosine similarity being one, can be effective as one neuron. For example, for $\alpha > 0$,
$$
a_1 \ReLU(\alpha \vw^{T} \vx) + a_2 \ReLU(\vw^{T} \vx) = (\alpha a_1 +a_2) \ReLU(\vw^{T} \vx).
$$
For the activation function $\tanh (x)$, the above reduction can not be rigorously correct, but only approximately.





\section{Condensation process during the training}
The condensation process during training plays a crucial role in understanding how over-parameterized neural networks can generalize effectively. Empirical observations suggest that the number of condensed clusters of neurons within the same layer typically increases monotonically as training progresses. Early in the nonlinear training phase, neurons tend to group into a small number of clusters. As training continues, the number of clusters expands, which aids in the network's ability to fit the data. Thus, the condensation phenomenon provides a mechanism for the growing complexity of the network’s outputs as training advances.

To illustrate this, consider the target function:
\begin{equation*}
    f(x) = - \sigma(x) + \sigma(2(x+0.3)) - \sigma(1.5(x-0.4)) + \sigma(0.5(x-0.8)),
\end{equation*}
where $\sigma(x) = \mathrm{ReLU}(x)$. The width of the hidden layer is $m=100$, and the learning rate is $0.1$. The parameters are initialized by $\mathcal{N}(0,\frac{1}{m^4})$. The training data is evenly sampled in $[-1,1]$.

\begin{figure}[htbp]
     \centering
     \begin{subfigure}[b]{0.275\textwidth}
         \centering
         \includegraphics[width=\textwidth]{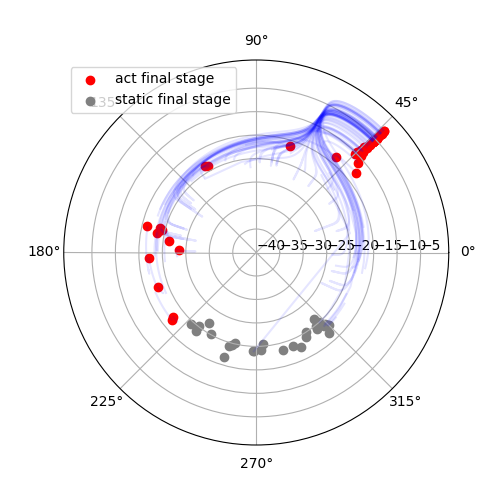}
         \caption{ epoch $=100$}
         \label{fig:ReLU_final_100}
     \end{subfigure}
     \hfill
     \begin{subfigure}[b]{0.275\textwidth}
         \centering
         \includegraphics[width=\textwidth]{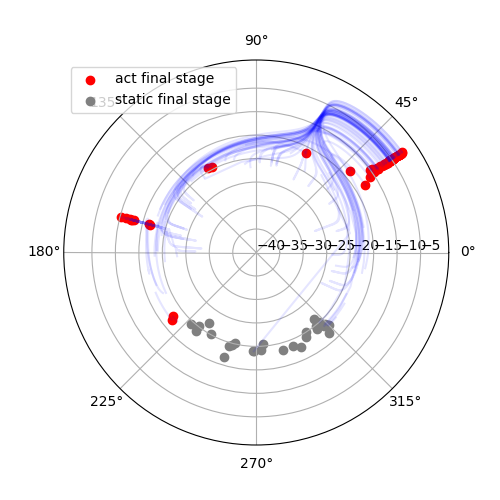}
         \caption{epoch $=1000$}
         \label{fig:ReLU_final_1000}
     \end{subfigure}
     \hfill
     \begin{subfigure}[b]{0.275\textwidth}
         \centering
         \includegraphics[width=\textwidth]{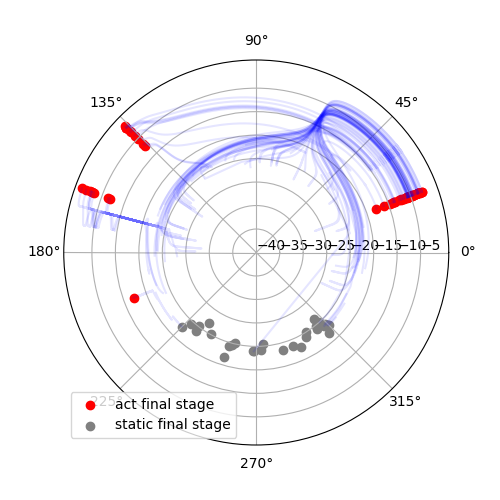}
         \caption{epoch $=5000$}
         \label{fig:ReLU_final_5000}
     \end{subfigure}

          \begin{subfigure}[b]{0.275\textwidth}
         \centering
         \includegraphics[width=\textwidth]{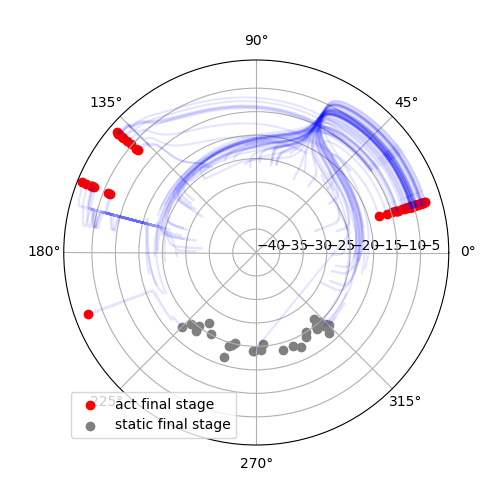}
         \caption{ epoch $=10000$}
         \label{fig:ReLU_final_10000}
     \end{subfigure}
     \hfill
     \begin{subfigure}[b]{0.275\textwidth}
         \centering
         \includegraphics[width=\textwidth]{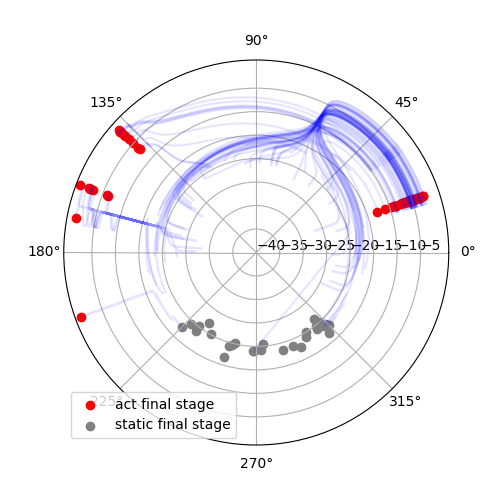}
         \caption{epoch $=12000$}
         \label{fig:ReLU_final_12000}
     \end{subfigure}
     \hfill
     \begin{subfigure}[b]{0.275\textwidth}
         \centering
         \includegraphics[width=\textwidth]{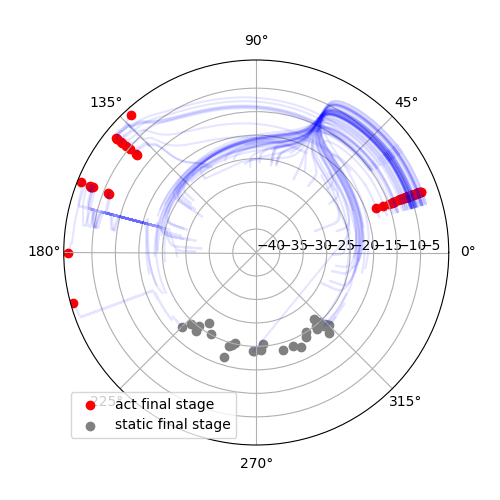}
         \caption{epoch $=100000$}
         \label{fig:ReLU_final_100000}
     \end{subfigure}
     
        \caption{The feature maps  $\{(\theta_k,A_k)\}_{k}$ of a two-layer ReLU neural network. The red dots and the gray dots are the features of the active and the static neurons respectively and the blue solid lines are the trajectories of the active neurons during the training. The epochs are described in subcaptions.}
        \label{fig:ReLU1dfinal}
\end{figure}

The features  $\{(\theta_k,A_k)\}_{k}$ during the training process are shown in Fig.~\ref{fig:ReLU1dfinal}. We observe that, as training progresses, the neurons in the network condense into a few isolated orientations, and the number of these orientations increases. A similar training process is shown in \cite{boursier2024early} on a piece-wise linear target function proposed in \cite{stewart2023regression}.

The presence of static neurons, which do not change their orientation during training, is attributed to the zero-gradient behavior induced by activation function $\mathrm{ReLU}(x)$. For all inputs, neurons always output zero; thus, no gradient during the training for these neurons.

\cite{yang2024effective} use an effective rank to measure the linear dependence among different neurons and found a staircase phenomenon, that is, the effective rank gradually increases in the training process, similar to a staircase curve. The staircase phenomenon is closely related to the condensation process. As the neurons condense into more and more groups, neurons in different groups are linearly independent, leading to the increase of effective rank. However, \cite{yang2024effective} do not notice that the scaling of initialization is a critical index to such non-linear phenomena, that is, the phase diagram of different training dynamics induced by initialization studied in \cite{luo2021phase}.


\section{More condensation experiments}\label{Sec:condense_exp}
This section will empirically give more examples from different network structures to show the condensation in training neural networks.
\subsection{Condensation in the synthetic data}

Consider a target function $f(x)=\mathrm{Tanh}(x)$. We also use a two-layer Tanh NN to fit the target function. The width of the hidden layer is $m=1000$, and the learning rate is $0.03$.  The training data is evenly sampled in $[-15,15]$.  The parameters are also initialized by $\mathcal{N}(0,(\frac{1}{m^{\gamma}})^2)$, where $\frac{1}{m^{\gamma}}$ is the standard deviation. 

\begin{figure}
    \centering
     \begin{subfigure}[b]{0.4\textwidth}
         \centering
         \includegraphics[width=\textwidth]{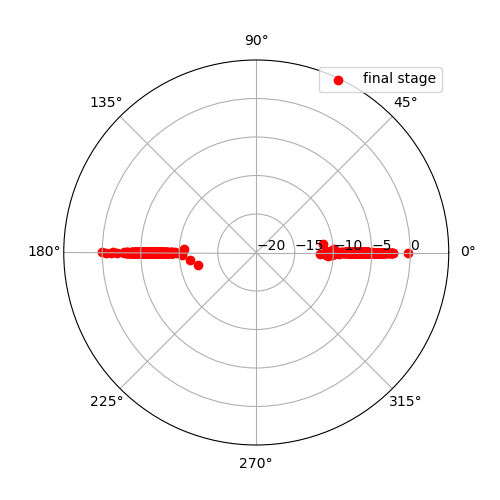}
         \caption{$\gamma=2$}
         \label{fig:Tanh_final_t2}
     \end{subfigure}
     \hfill
     \begin{subfigure}[b]{0.4\textwidth}
         \centering
         \includegraphics[width=\textwidth]{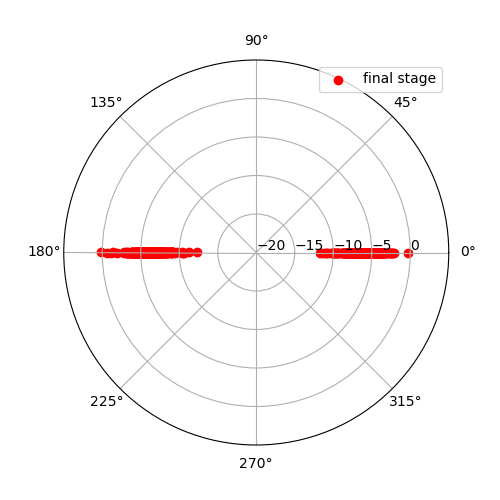}
         \caption{$\gamma=3$}
         \label{fig:Tanh_final_t6}
     \end{subfigure}
    \caption{The feature map of two-layer Tanh neural networks. The red dots are the features of neurons at the terminal stage. The initialization scales are indicated in the subcaptions.}
    \label{fig:Tanh_final_30000}
\end{figure}

Fig.~\ref{fig:Tanh_final_30000} shows the terminal stage of two-layer Tanh NNs with different initializations.  The neurons condense to a pair of opposite directions when the training converges. And as the initializations become smaller, the neurons become more condensed. 

\subsection{Condensation in the CNNs}
We trained a convolutional neural network with only one convolutional layer using the MNIST dataset (a commonly used small image dataset) and cross-entropy loss as the loss function. 

\begin{figure}[h!]
     \centering
     \subfloat[Loss]{\includegraphics[width=0.33\textwidth]{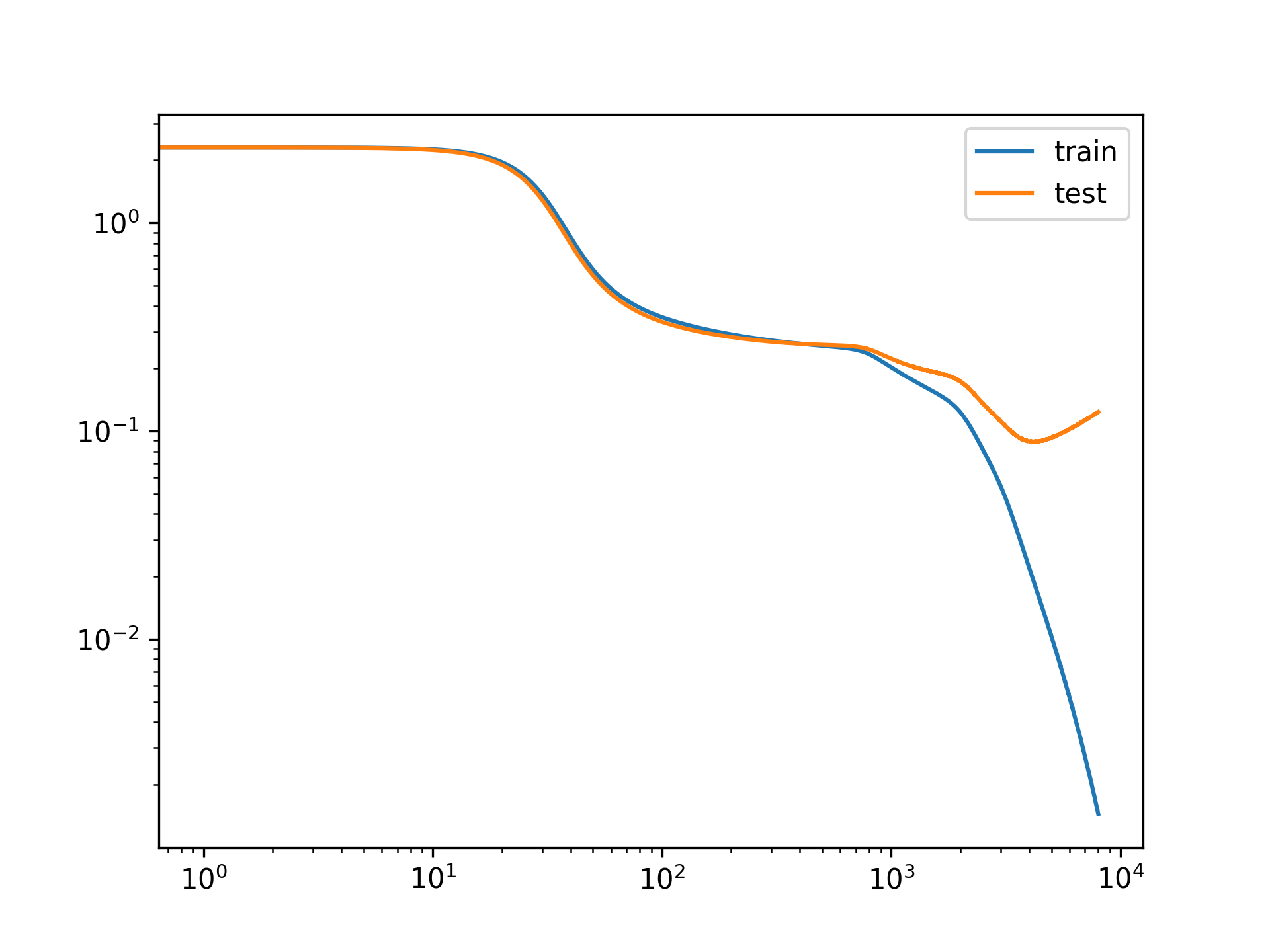}}
     \subfloat[initial weight]{\includegraphics[width=0.33\textwidth]{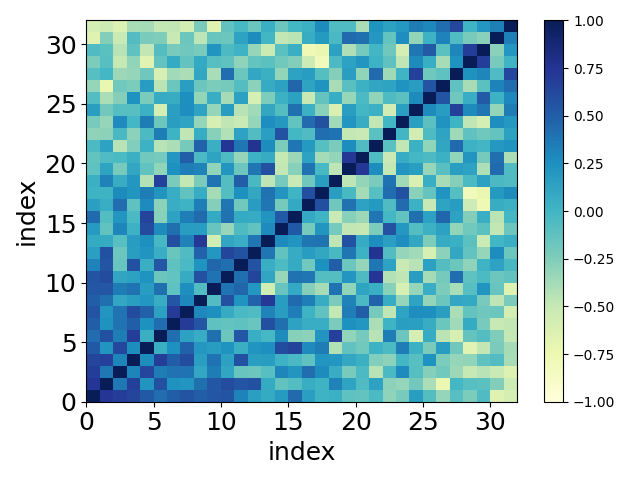}}
    \subfloat[initial output]{\includegraphics[width=0.33\textwidth]{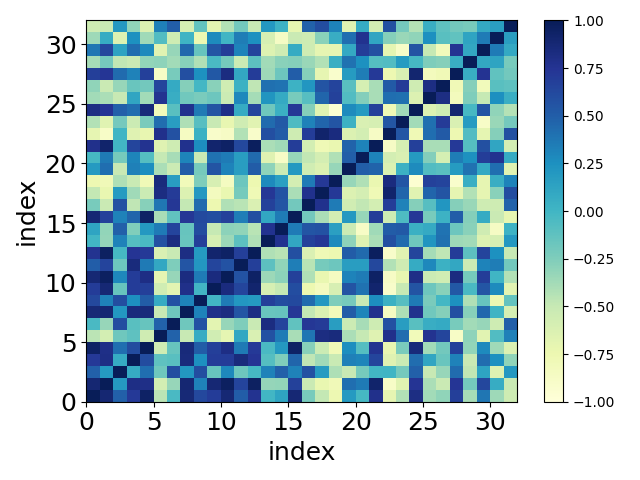}}

     \hfill
    \subfloat[Accuracy]{\includegraphics[width=0.33\textwidth]{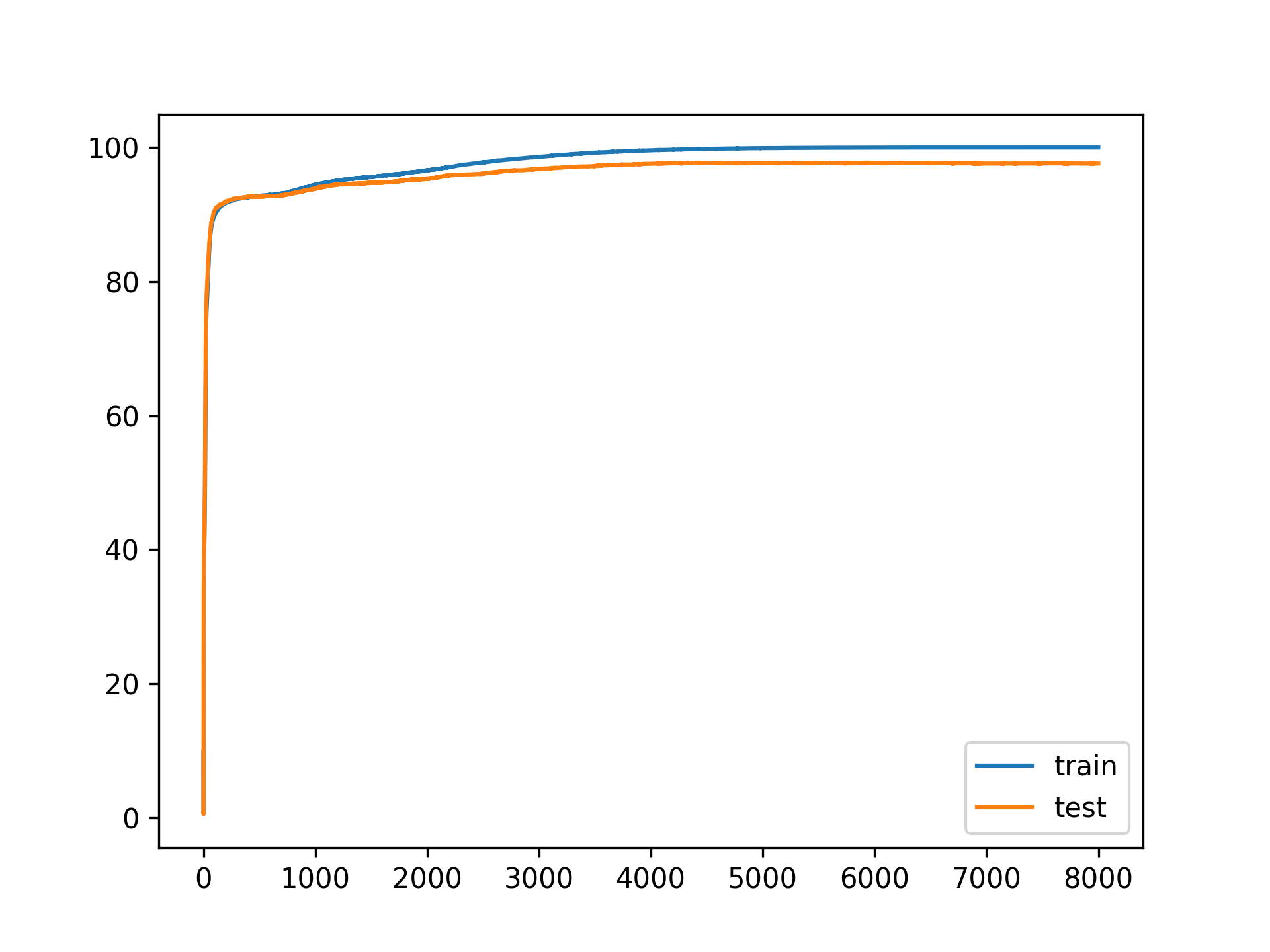}}
    \subfloat[final weight]{\includegraphics[width=0.33\textwidth]{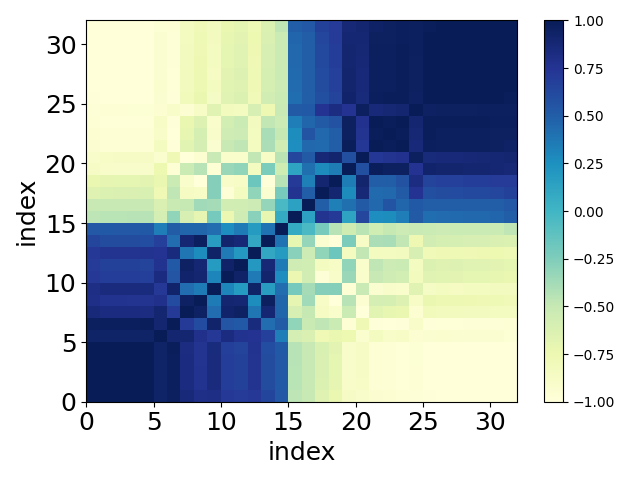}}
    \subfloat[final output]{\includegraphics[width=0.33\textwidth]{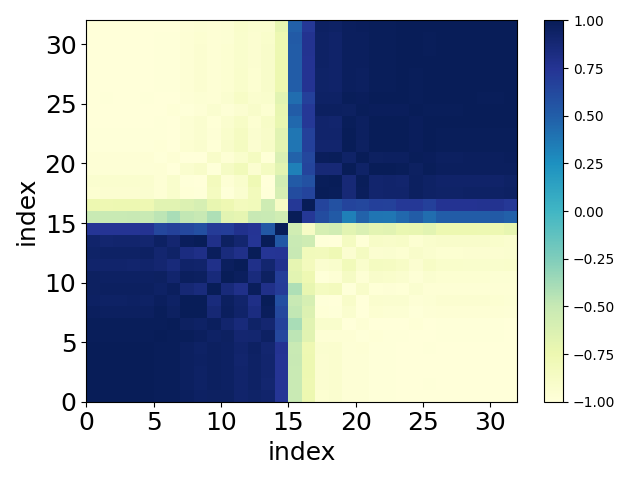}}

     \caption{Small initialization (convolutional and fully connected layers initially follow $\mathcal{N}(0,96^{-8})$) for single-layer CNN training in its final stage of convergence. The activation function is $\rm{tanh}(x)$. If neurons are in the same dark blue block, then $D(\vu,\bm{v})\sim 1$ (in beige blocks, $D(\vu,\bm{v})\sim -1$), indicating that their input weight directions are the same (opposite). Colors represent $D(\vu,\bm{v})$ of two convolution kernels, with indices shown on the horizontal and vertical axes respectively. The training set is MNIST. The output layer uses softmax, the loss function is cross-entropy, and the optimizer is Adam with full batch training. Convolution kernel size $m=3$, learning rate $=2 \times 10^{-4}$. Training continues until $100\%$ accuracy is achieved on the training set, at this point, the test set accuracy is $97.62\%$. }
        
        \label{fig:tanhCNNfinal}

\end{figure}

Fig.~\ref{fig:tanhCNNfinal}(a) and (d) show the loss and accuracy during the training process, respectively. Fig.~\ref{fig:tanhCNNfinal}(b) and (e) display the cosine similarity heatmaps of the convolution kernels at the beginning of training and when the training accuracy reaches $100\%$, respectively. The convolutional layer has $32$ channels with a kernel size of $3\times 3$, resulting in cosine similarities between $32$ different $9$-dimensional weight vectors. 


Fig.~\ref{fig:tanhCNNfinal}(c) and (f) show the cosine similarities of the neural network output vectors. These vectors were obtained by passing a combined dataset of $70,000$ data points from both the training and test sets through the convolutional layer, resulting in a $4$-dimensional tensor of size $70000\times 32\times 28\times 28$. We fixed the second dimension and flattened the remaining dimensions. This allowed us to compute the cosine similarities between $32$ vectors, each of size $70000\times28\times28$.


The figures reveal two key observations. First, at initialization, no clustering relationship exists between the vectors. However, after training is completed, block-like structures emerge both in the convolutional layer and in the data processed by the convolutional layer, indicating the presence of the condensation phenomenon. The vectors tend to converge in two opposite directions. Second, the block structure in Fig.~\ref{fig:tanhCNNfinal}(f) is more pronounced than in Fig.~\ref{fig:tanhCNNfinal}(e), suggesting that the degree of condensation in the output of the convolutional layer is more pronounced than weights in the final-stage.

\subsection{Condensation in the residual CNN}

The condensation phenomenon also occurs in residual neural networks. We use the deep learning network model ResNet18 as an example to demonstrate the condensation phenomenon during its training process.

ResNet18 is a convolutional neural network applied to visual tasks, excelling in processing images. The network consists of $18$ main learnable parameter layers ($17$ convolutional layers, $1$ linear layer), batch normalization layers, pooling layers, etc. These layers are organized in a specific structure called residual blocks. Although ResNet18 is relatively small in scale among deep learning models, it can achieve a top-1 accuracy of $73.16\%$ and a top-5 accuracy of $91.03\%$ on the ImageNet dataset\footnote{source: \url{https://huggingface.co/timm/resnet18.a1_in1k}}.

In residual neural networks, we handle convolutional kernels similarly to convolutional neural networks, with the only difference being that multi-channel convolutional kernels need to be flattened across both channels and kernel dimensions. For the neural network output, we randomly select $256$ training images and $256$ test images to form a batch of $512$ images and observe the condensation among vectors in this batch using a process similar to that used in convolutional neural networks.

\begin{figure}[h!]
	\centering
	\subfloat[]{
		\includegraphics[width=0.45\textwidth]{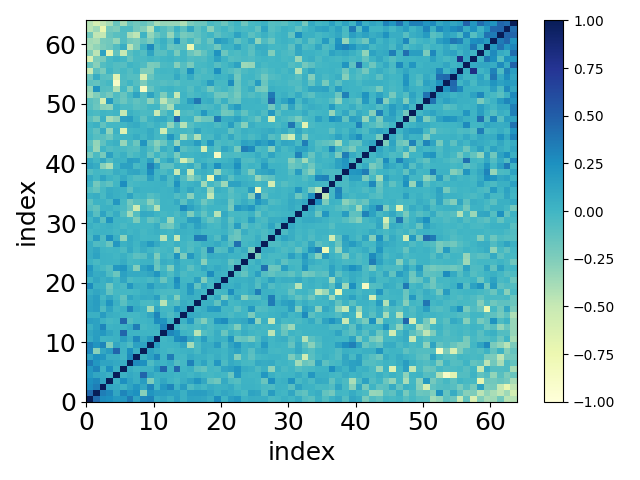}
	}
	\subfloat[]{
		\includegraphics[width=0.45\textwidth]{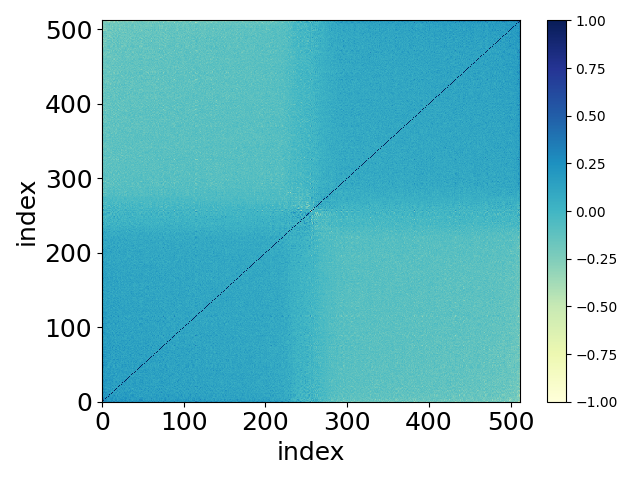}
	}
 \\
 	\subfloat[]{
		\includegraphics[width=0.45\textwidth]{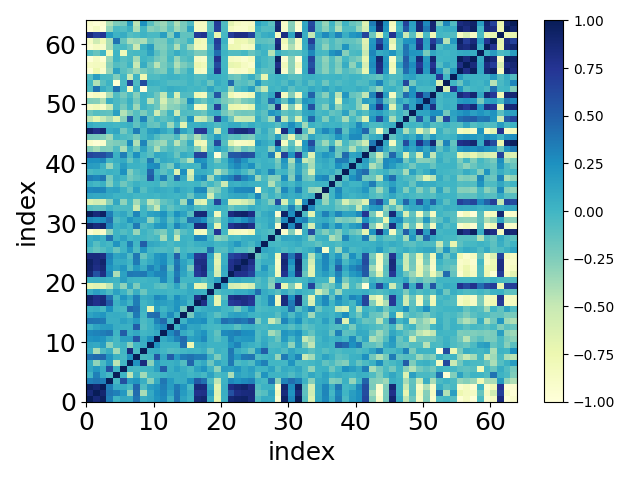}
	}
	\subfloat[]{
		\includegraphics[width=0.45\textwidth]{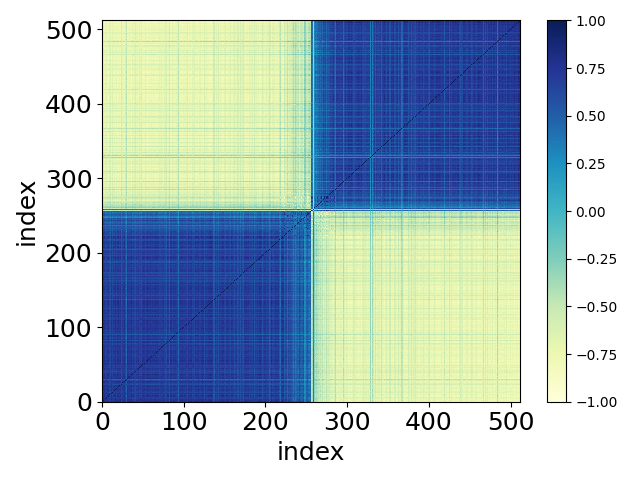}
	}

	\caption{Condensation phenomenon in a ResNet-18 model pre-trained on ImageNet. (a) and (b) show weights from the first and the last convolutional layers of ResNet-18 respectively, and (c) and (d) are the corresponding outputs.}
	\label{fig:pretrained_heatmap}
\end{figure}

As shown in Fig.~\ref{fig:pretrained_heatmap}(b) and (d), both the weights and outputs of the last convolutional layer exhibit condensation, while the weights and outputs of the first layer (as shown in Fig.~\ref{fig:pretrained_heatmap}(a) and (c)) do not demonstrate such pronounced condensation. This experiment shows that different layers would have different degrees of condensation.

\section{Phase diagram: when condensation happens}
Empirically, we have found that in non-linear training regime, condensation is a very common phenomenon. In Ref. \cite{luo2021phase}, to characterize the non-linear and linear regimes, we consider a two-layer NN with $m$ hidden neurons
\begin{equation}\label{eq: 2LNN}
    f^{\alpha}_{\vtheta}(\vx) = \frac{1}{\alpha}\sum_{k=1}^{m}a_k\sigma(\vw_k^{\T}\vx),
\end{equation}
where $\vx\in\sR^{d}$, $\alpha$ is the scaling factor, $\vtheta=\mathrm{vec}(\vtheta_a,\vtheta_{\vw})$ with $\vtheta_a=\mathrm{vec}(\{a_k\}_{k=1}^{m})$, $\vtheta_{\vw}=\mathrm{vec}(\{\vw_k\}_{k=1}^{m})$ is the set of parameters initialized by $a_k^0\sim N(0, \beta_1^2)$, $\vw_k^0\sim N(0, \beta_2^2 \mI_d)$. The bias term $b_k$ can be incorporated by expanding $\vx$ and $\vw_k$ to $(\vx^\T,1)^\T$ and $(\vw_k^\T,b_k)^\T$. We consider  the infinite-width limit $m\to\infty$.

The linear regime refers to a dynamic regime that the model can be approximated by the first-order Taylor expansion at the initial parameter point, i.e., 
\begin{equation}\label{eq: 2LNN-linear}
    f^{\alpha}_{\vtheta(t)}(\vx) \approx f^{\alpha}_{\vtheta(0)}(\vx) + \nabla_{\vtheta} f^{\alpha}_{\vtheta(0)}(\vx)\cdot (\vtheta(t)-\vtheta(0)),
\end{equation}
where $\vtheta(t)$ is the parameter set at training time $t$. Therefore, to characterize the linear/non-linear regime, the key is the change of $\vtheta_{\vw}$ during the training. If it changes very slightly, then, the first-order Taylor expansion can be approximated held, i.e., linear regime, otherwise, non-linear regime. A key quantity is defined as:
\begin{equation}
    \mathrm{RD}(\vtheta_{\vw}(t))=\frac{\norm{\theta_{\vw}(t)-\theta_{\vw}(0)}_{2}}{\norm{\theta_{\vw}(0)}_{2}}.
\end{equation}

\begin{figure}
    \centering
    \includegraphics[width=\textwidth]{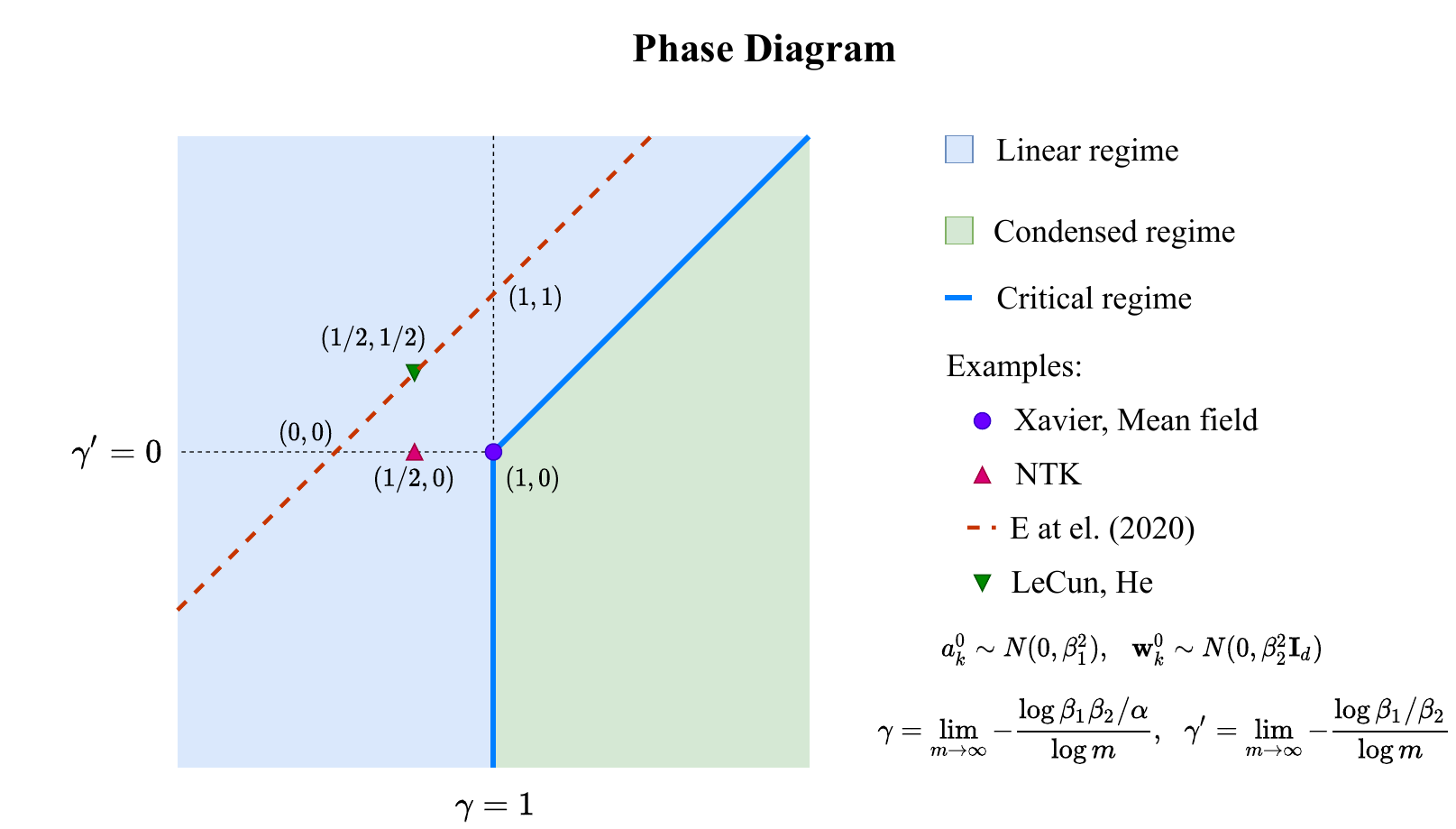}
    \caption{Phase diagram of two-layer ReLU NNs at infinite-width limit. The marked examples are studied in existing literature. Table is from Ref. \cite{luo2021phase}. }
    \label{fig:phase-diagram}
\end{figure}

Through appropriate rescaling and normalization of the gradient flow dynamics, which accounts for the dynamical similarity up to a time scaling, we arrive at two independent coordinates
\begin{equation}
    \gamma=\lim\limits_{m\to\infty}-\frac{\log\beta_1\beta_2/\alpha}{\log m}, \quad \gamma'=\lim\limits_{m\to\infty}-\frac{\log\beta_1/\beta_2}{\log m}.
\end{equation}
The resulting phase diagram is shown in Fig.~\ref{fig:phase-diagram}, which can be rigorously characterized by the following two theorems.
\begin{thm}
    [Informal statement \cite{luo2021phase}] If $\gamma<1$ or $\gamma'>\gamma-1$, then with a high probability over the choice of $\vtheta^0$, we have
    \begin{equation}
        \lim_{m\to+\infty}\sup\limits_{t\in[0,+\infty)}\mathrm{RD}(\vtheta_{\vw}(t))=0.
    \end{equation}
\end{thm}
\begin{thm}
    [Informal statement \cite{luo2021phase}] If $\gamma>1$ and $\gamma'<\gamma-1$, then with a high probability over the choice of $\vtheta^0$, we have
    \begin{equation}
        \lim_{m\to+\infty} \sup\limits_{t\in[0,+\infty)}\mathrm{RD}(\vtheta_{\vw}(t))=+\infty.
    \end{equation}
\end{thm}
For the non-linear regime, we find that condensation is a unique feature, therefore, we name it condensation regime. For three-layer ReLU neural networks, we found similar phase diagrams for the dynamics of each layer \cite{zhou2022empirical}.


The study of phase diagrams provides valuable insights into how to appropriately tune parameter initialization when scaling up network sizes, which we elaborate in the next section.

\section{Initialization scaling exponent as a hyperparameter}\label{Sec:gamma_hyperparameter}

The study of phase diagrams provides valuable insights into how to appropriately tune parameter initialization when scaling up network sizes. A commonly used initialization method involves sampling the parameters from a Gaussian distribution $\fN(0,(\frac{1}{m^{\gamma}})^2)$ (or a uniform $[-\frac{1}{m^{\gamma}},\frac{1}{m^{\gamma}}]$ distribution), where $\frac{1}{m^{\gamma}}$ is the standard deviation and $m$ represents the input dimension or the average of the input and output dimensions. When scaling up network sizes, to maintain similar dynamic behavior, it is crucial not to fix the initialization standard deviation, but rather to keep $\gamma$ fixed. The exponent $\gamma$ directly determines the training regime --- whether the network operates in the linear regime or the condensation regime --- and thus fundamentally shapes the optimization landscape and generalization behavior.

In current mainstream deep learning frameworks such as PyTorch, TensorFlow, JAX, and Megatron-LM, the initialization of network parameters is typically specified through named schemes. For example, PyTorch provides \texttt{kaiming\_uniform}, \texttt{kaiming\_normal}, \texttt{xavier\_uniform}, and \texttt{xavier\_normal}, each corresponding to a specific choice of the scaling exponent $\gamma$ (e.g., $\gamma=1/2$ for Kaiming and Xavier family). While these built-in schemes are convenient, they implicitly fix $\gamma$ and offer no straightforward mechanism for users to explore alternative values. In practice, switching from, say, $\gamma=1/2$ to $\gamma=3/4$ requires manually computing the desired standard deviation and calling low-level initialization functions, which is error-prone and obscures the underlying intent.

We propose that the initialization scaling exponent $\gamma$ should be explicitly exposed as a first-class hyperparameter in deep learning frameworks. Just as learning rate, weight decay, and momentum are standard tunable parameters in an optimizer, $\gamma$ governs a qualitatively different and equally important aspect of training: the dynamical regime of the network. A practical interface could take the form of an additional argument in the initialization API, e.g., \texttt{nn.init.kaiming\_normal\_(tensor, gamma=0.5)}, or a global configuration option that adjusts all layer initializations consistently.

There are several concrete benefits to this proposal. First, it enables reproducible exploration of training regimes. Researchers and practitioners can systematically sweep over $\gamma$ values to identify whether their architecture and data benefit from condensation or linear dynamics, without rewriting initialization code. Second, it ensures consistency when scaling model sizes. As models grow wider or deeper, the phase diagram tells us that maintaining the same $\gamma$ preserves the dynamical regime, whereas naively preserving the standard deviation (a fixed $\sigma$) can inadvertently shift the network into a different regime, leading to unexpected training failures. Third, framework-level support for $\gamma$ facilitates the transfer of theoretical insights into practice. The phase diagram framework provides rigorous guidance on how $\gamma$ affects training, but this guidance can only be widely adopted if the relevant parameter is easy to control in code.

We encourage the deep learning community and framework developers to consider incorporating $\gamma$ as an explicit, tunable hyperparameter. This small interface change would bridge the gap between the theoretical understanding of training dynamics and everyday deep learning practice, enabling more principled initialization strategies as models continue to scale.

\section{Mechanisms underlying condensation}
The condensation phenomenon is not yet fully understood. However, a series of studies have provided valuable insights into the mechanisms underlying condensation. In this review, we provide an overview of three perspectives: initial condensation through training dynamics, the implicit regularization effect of dropout training, and the structure of critical points in the loss landscape.

\subsection{Initial condensation}\label{Sec:initial_condense}
Neurons within the same layer exhibit an important symmetry property: swapping the indices of any two neurons does not affect the system's behavior. When we describe the dynamics of a neuron, the dynamics of any other neuron within the same layer can be obtained by simply swapping their indices. Formally, the dynamics of all neurons within the same layer follow the same ordinary differential equation (ODE). If this ODE has a finite number of stable points, and the number of neurons exceeds the number of stable points, many neurons will evolve towards the same stable points.

Quantifying this dynamic process precisely is challenging due to the nonlinearity of the training process. However, in certain specific scenarios, this analysis can be further developed.

For gradient descent training, small initialization plays a crucial role in influencing condensation. The analysis can be approached by taking the limit as the initialization approaches zero. In this case, the output of the neural network simplifies. Two scenarios are studied: one for activation functions that are differentiable at the origin, and the other for the ReLU activation function.

\begin{figure}[h]
	\centering
	\subfloat[$\mathrm{Tanh}$, $p=1$]{\includegraphics[width=0.44\textwidth]{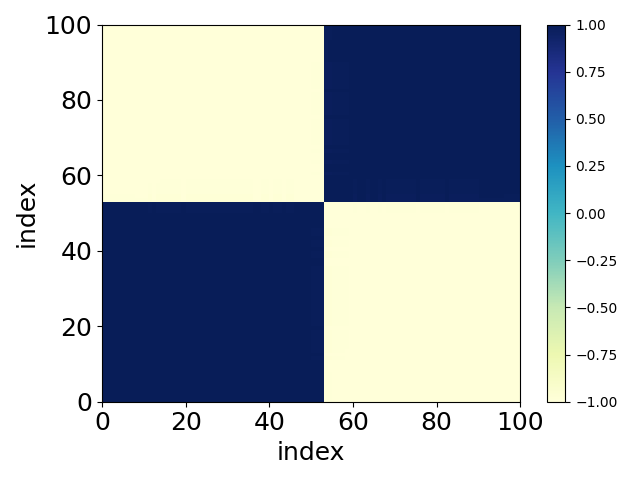}}
	\subfloat[$\mathrm{xTanh}$, $p=2$]{\includegraphics[width=0.44\textwidth]{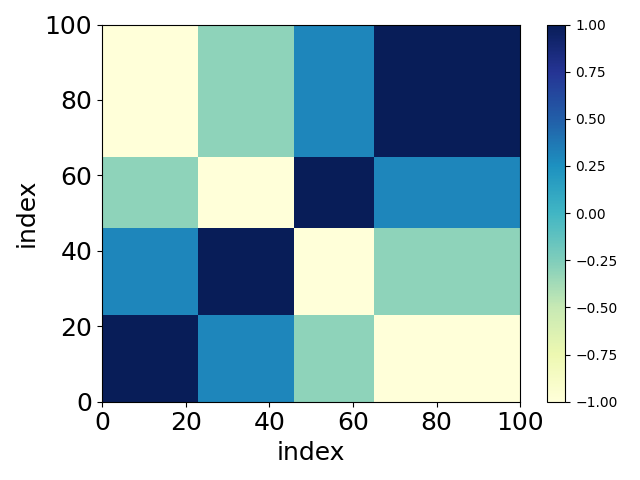}}

  \caption{The heatmap of the cosine similarity of neurons of two-layer NNs at the initial training stage. The activation functions and their corresponding multiplicities are specified in the subcaptions. The target function is $\sin(x)$. The parameters of all layers are initialized following $\mathcal{N}(0,\frac{1}{m^4})$. The optimizer is Adam. The width $m=100$ and the learning rate is $0.0005$. The plot epochs are $100$ and $200$ respectively. \label{fig:Initial_feature}}
\end{figure}

For the first case, the network output can be approximated by the leading-order term of the activation function, where the leading order is denoted as $p$.
\begin{defi}[multiplicity $p$ \cite{zhou2022towards}]\label{defi:multiplicity}
    Suppose that $\sigma(x)$ satisfies the following condition, there exists a $p\in \sN^{*}$, such that the $s$-th order derivative $\sigma^{(s)}(0)=0$ for $s=1,2,\cdots,p-1$, and $\sigma^{(p)}(0)\neq 0$, then we say $\sigma$ has multiplicity $p$.
\end{defi}
Experiments in \cite{zhou2022towards} suggest that the maximum number of condensed directions for input weights is no greater than $2p$. Additionally, theoretical analysis is provided for the case of $p=1$, as well as for any $p$ with one-dimensional input. For the case of $p=1$, \cite{chen2023phase} further estimates the time required for initial condensation in two-layer NNs. The following example illustrates how the activation function can influence the initial condensed directions. As is shown in Fig.~\ref{fig:Initial_feature}, when employing $\mathrm{Tanh}$ as the activation, there are a pair of opposite condensed directions. When the activation function is $\mathrm{xTanh}$, there are two pairs of opposite condensed directions. 

In the case of $p=1$, several works investigate different scenarios. \cite{chen2024dynamics} shows that three layer NNs will have condensed solutions at the initial stage with some assumptions. \cite{zhou2023understanding} analyzes the initial condensation of two-layer convolutional NNs. \cite{chen2024analyzing} analyzes the subsequent loss descent and the second loss plateau after the initial condensation stage.


For the second case, \cite{maennel2018gradient} shows that in the limit of infinitesimal initial weights and learning rate, two-layer ReLU NN will first align at a discrete set of possible directions before the loss descent. \cite{phuong2021the} analyzes a more concrete setting on the orthogonally separable data and the neurons will asymptotically converge to two neurons: the positive max-margin vector and the negative max-margin vector. \cite{boursier2022gradient} investigates the time of the early alignment stage when the data forms an orthonormal family. \cite{chistikov2023learning} observes that when using a two layer ReLU network to learn a target function of one neuron with correlated inputs, the neurons will first align and will not separate during training. \cite{wang2024understanding} estimates the time of the early alignment phase in the binary classification problem of  effectively two data points, which are separated by small angles, and \cite{min2024early} looses the data assumption to that the data are positively correlated when they have the same labels.  \cite{boursier2024early} demonstrates a quantitative analysis of the initial condensation of both regression and classification and general datasets in two layer NNs. They also give an example that the initial condensation will do harm to the final convergence with the initialization that $|a_j|\geq||w_j||$.  \cite{kumar2024directional,kumar2024early} extends the analysis of early alignment to homogeneous neural networks, with \cite{kumar2024directional} exploring alignment dynamics that near saddle points beyond initialization on two-homogeneous NNs. \cite{lyu2021gradient} demonstrates that a two-layer leaky ReLU NN with linear separable and symmetric data will align in the first phase and finally reach a global-max-margin linear classifier.

\subsection{Embedding principle}
The condensation phenomenon suggests that a large network in the condensed state is effectively equivalent to a much smaller network. This raises two important questions: Why not simply train a smaller network to save computational cost? What are the similarities and differences between a large network and a small network that share the same output function?

To explore these questions, we conduct experiments using two-layer ReLU networks with different widths to fit the same one-dimensional target function. 

\begin{figure}
    \centering
    \includegraphics[width=0.75\textwidth]{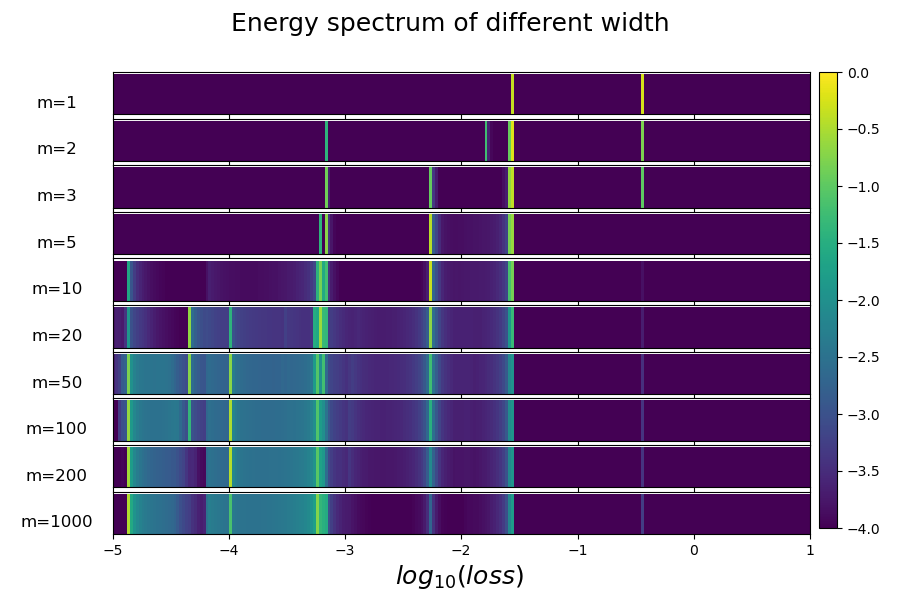}
    \caption{The loss distribution during the training among two-layer ReLU NNs with different widths. Each row is the probability of loss in $50$ trials at the width of $m$ where each trial processes $10^5$ epochs. The probability is shown on the log scale. The experiment setting is the same as Fig.~\ref{fig:ReLU1dfinal}.}
    \label{fig:energy_bar}
\end{figure}

For each network width $m$, we train the network for 50 trials with different random seeds, resulting in 50 training loss curves. For each loss bin interval, we sum the number of training epochs during which the loss values fall within that interval across all trials. This sum is then normalized by the total number of epochs to obtain the frequency for that loss interval, which is represented by the color in the corresponding row of Fig.~\ref{fig:energy_bar}.

The loss that exhibits a bright bar in the figure indicates that the training trajectory remains close to this loss value for a significant number of epochs. Given that the gradient is likely small, the trajectory can persist at this point for many epochs, suggesting that such a point is highly likely to be a critical point. Comparing the loss distributions across different network widths, we observe that networks of varying widths tend to encounter similar critical points. However, as the network width increases, there is a greater likelihood that the training losses will remain at lower values. This suggests a difference in behavior, namely, that larger networks may find it easier to escape saddle points.

To understand the similarities and differences among networks with varying widths, \cite{zhang2021embedding} introduced an \textbf{embedding principle}, which states that the loss landscape of any neural network ``contains'' all critical points of all narrower networks. Similar ideas are also studied in \cite{fukumizu2000local, fukumizu2019semi,csimcsek2021geometry}. Specifically, for a narrow network at a critical point, if a neuron is split into two neurons in the following manner: the new neurons have the same input weights as the original one, and the sum of the output weights of the two new neurons to a subsequent neuron equals the output weight of the original neuron to that subsequent neuron, then the wider network will also be at a critical point. This explains the similarities shared by networks of various widths. It is important to note that the wider network can be regarded as in a condensed state.

Furthermore, \cite{zhang2022embedding} reveals that when embedding a critical point from a narrow neural network into a wider network, the numbers of positive, zero, and negative eigenvalues of the Hessian at the critical point are non-decreasing. This theorem suggests that a local minimum may transition into a saddle point due to the potential increase in negative eigenvalues during the embedding process. Additionally, the growth in negative eigenvalues facilitates easier escape from saddle points during training. Simultaneously, the increase in the number of zero eigenvalues makes it more likely for training trajectories to be attracted to that critical point.

The embedding principle is an intrinsic property of networks with a layered structure, independent of the target function, loss function, or optimization method. It provides a rationale for the emergence of condensation from the perspective of the loss landscape.

\subsection{Dropout facilitates the condensation}
\begin{figure}[h]
	\centering
	\subfloat[$p=1$, output]{\includegraphics[width=0.24\textwidth]{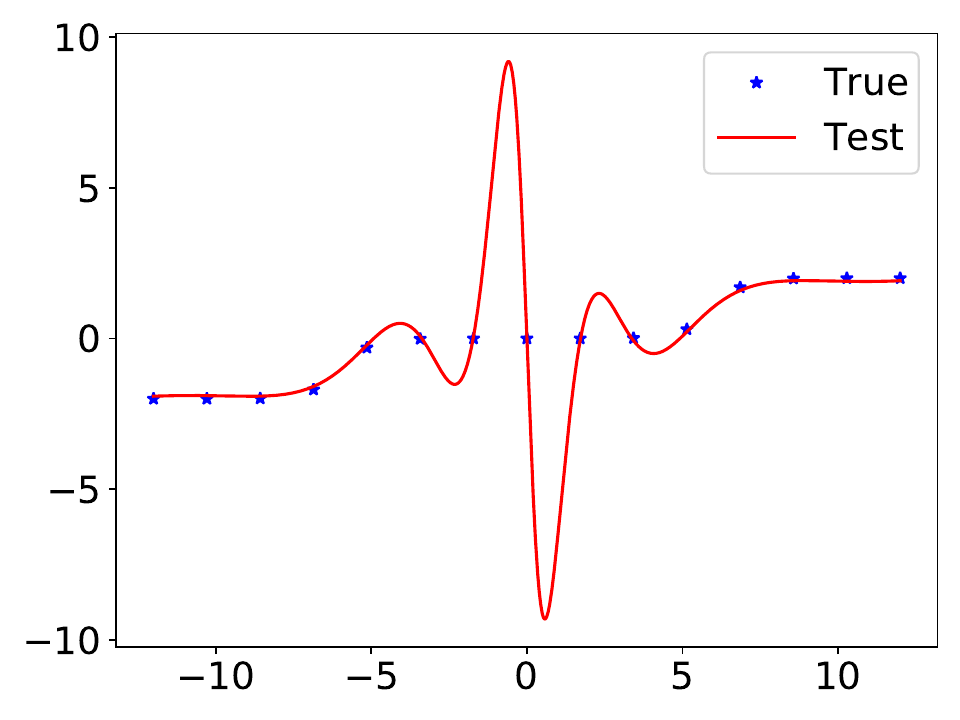}}
	\subfloat[$p=0.9$, output]{\includegraphics[width=0.24\textwidth]{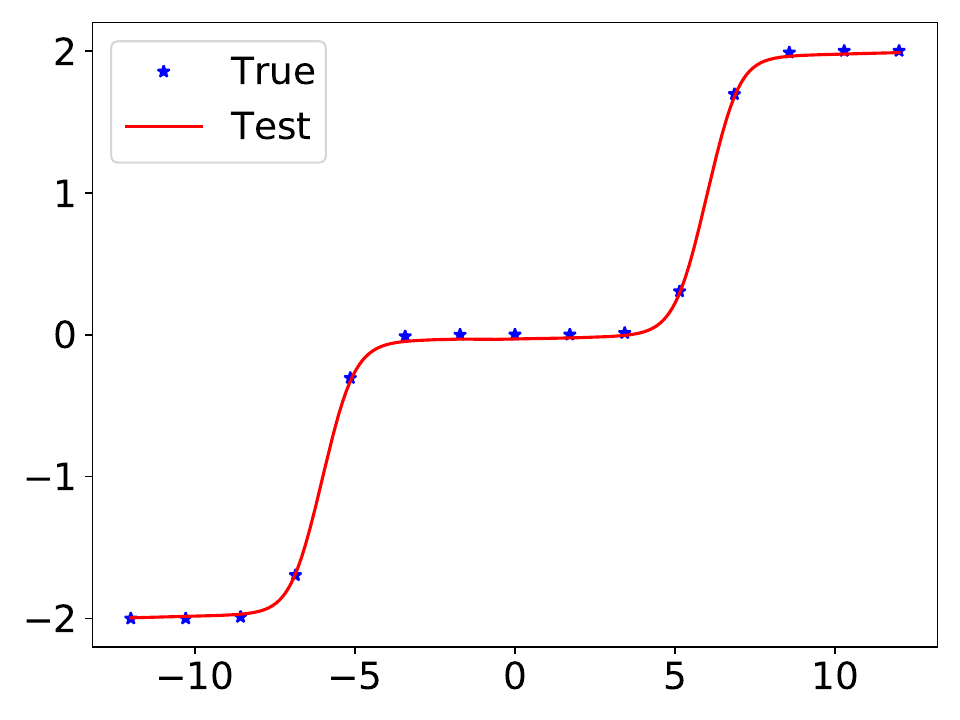}}
	\subfloat[$p=1$, feature]{\includegraphics[width=0.24\textwidth]{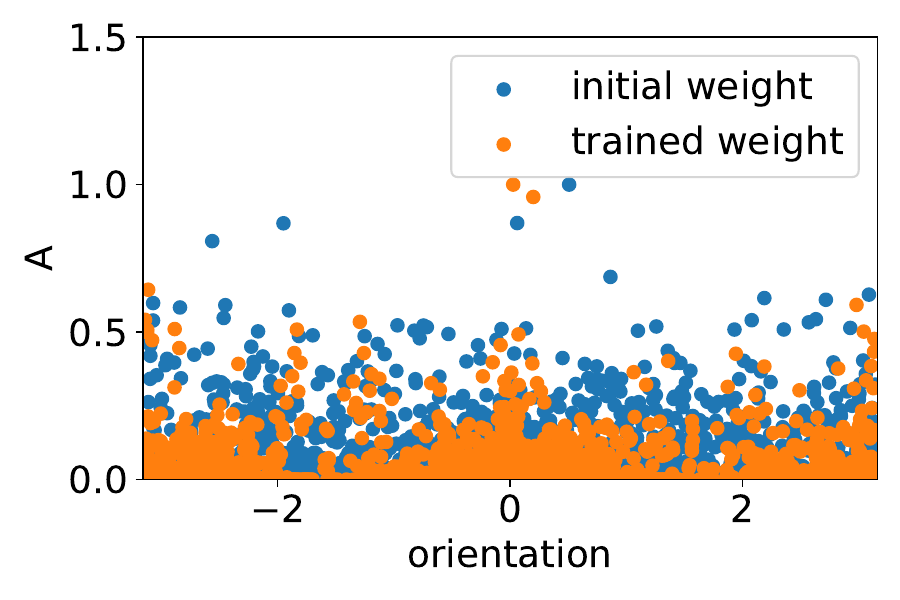}}
	\subfloat[$p=0.9$, feature]{\includegraphics[width=0.24\textwidth]{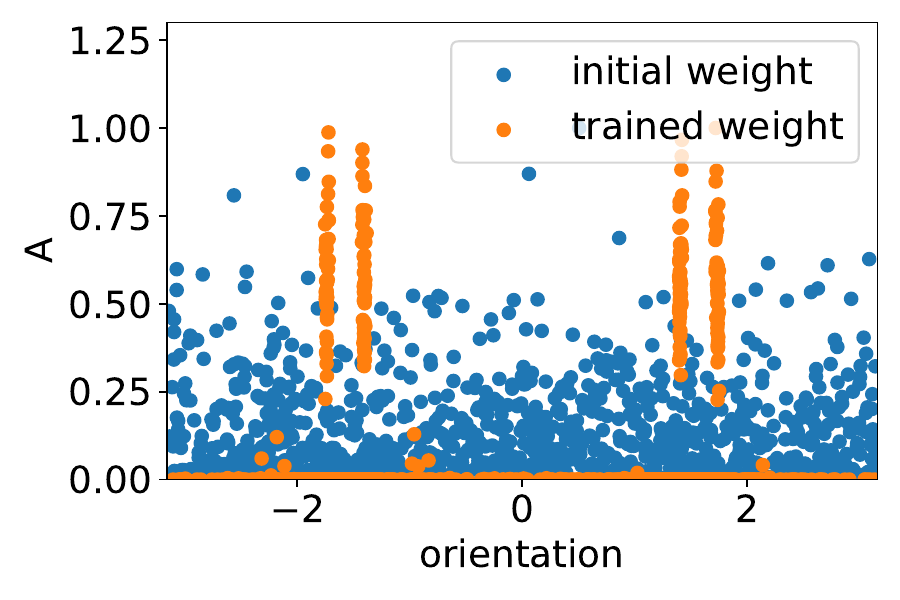}}\\
	\subfloat[$p=1$, output]{\includegraphics[width=0.24\textwidth]{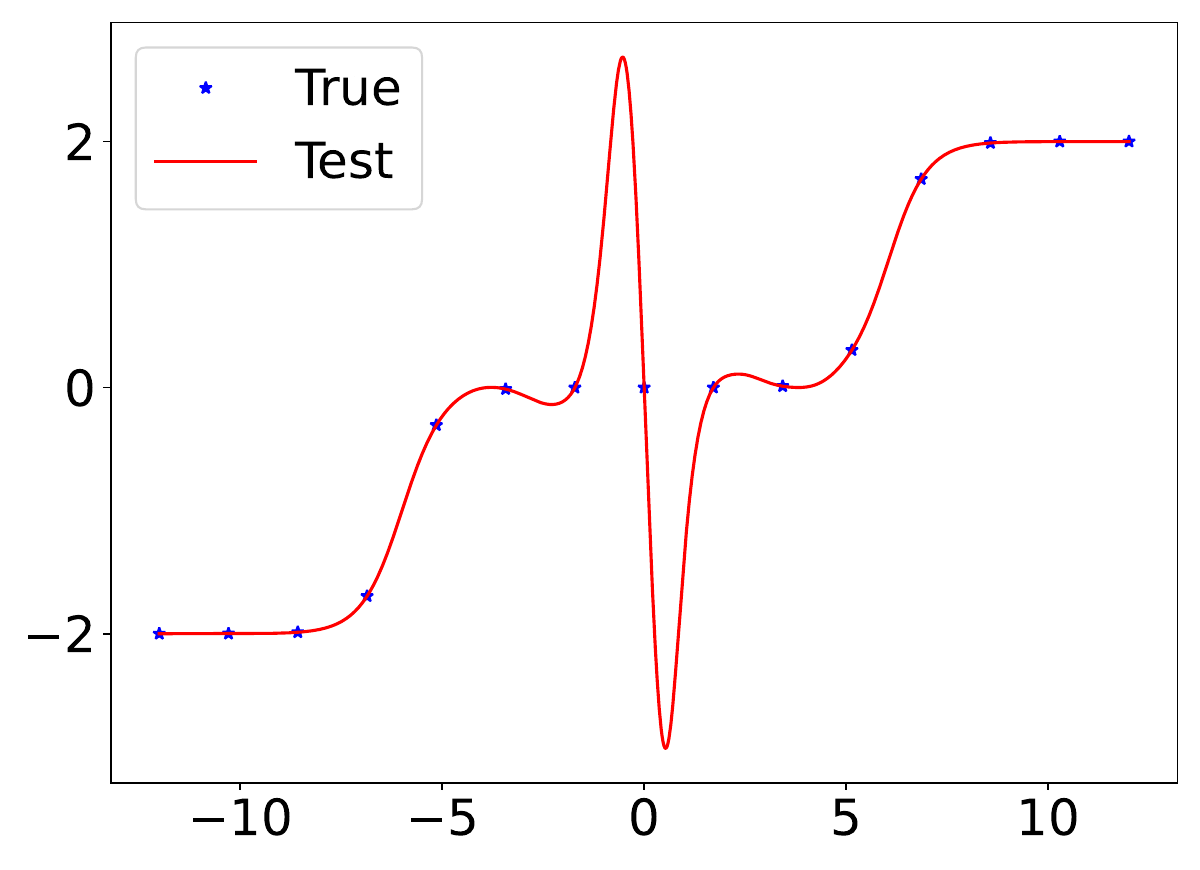}}
	\subfloat[$p=0.9$, output]{\includegraphics[width=0.24\textwidth]{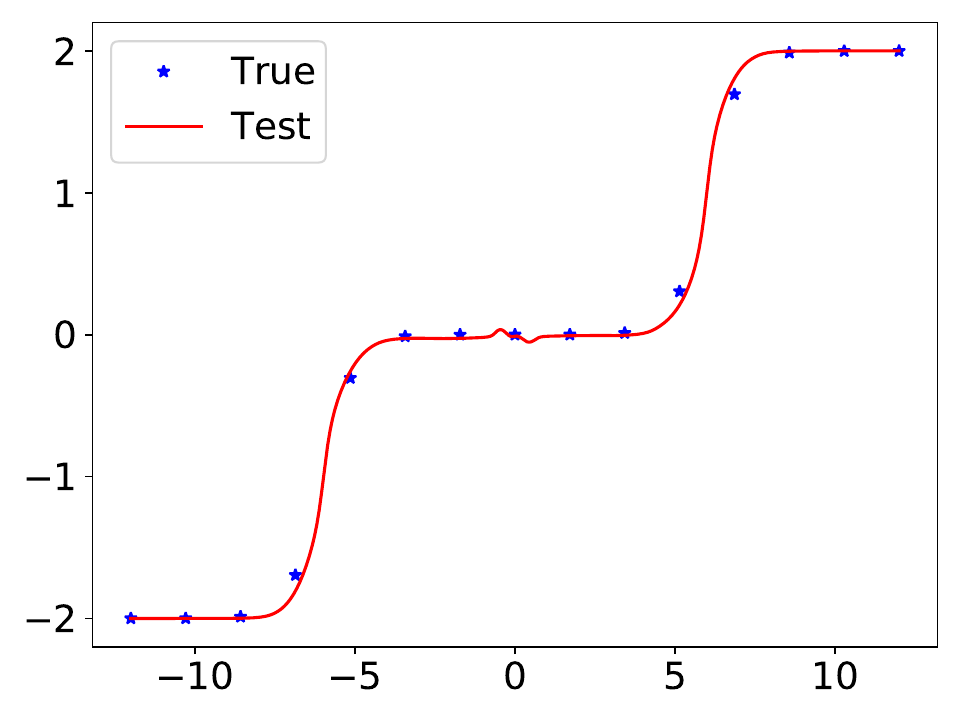}}
	\subfloat[$p=1$, feature]{\includegraphics[width=0.24\textwidth]{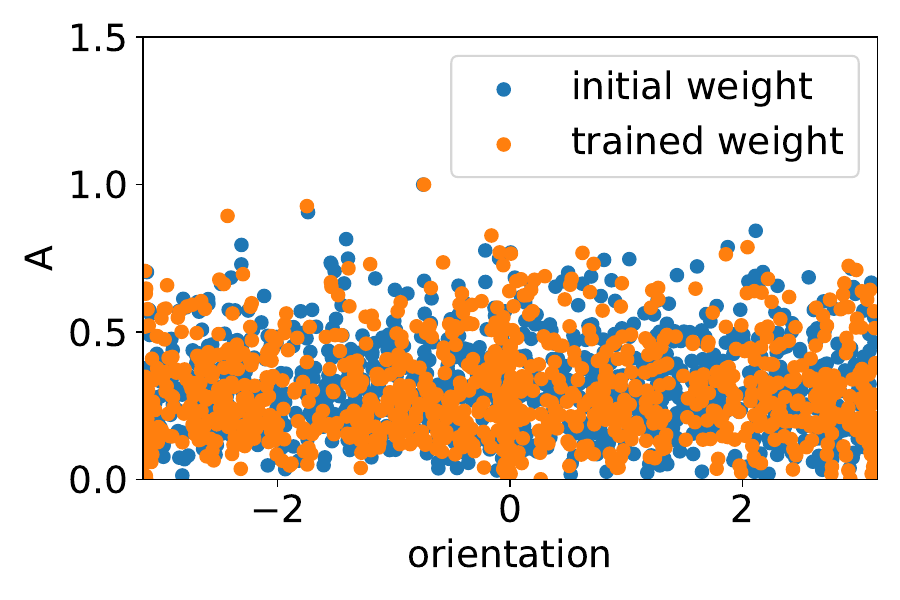}}
	\subfloat[$p=0.9$, feature]{\includegraphics[width=0.24\textwidth]{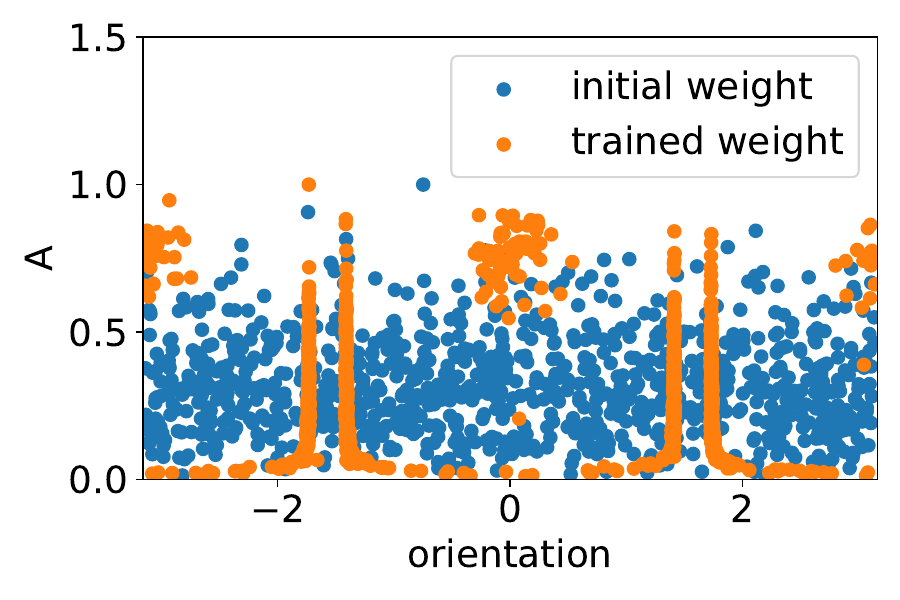}}

  \caption{ Tanh NNs outputs and features under different dropout rates. The width of the hidden layers is $1000$, and the learning rate for different experiments is $1\times10^{-3}$. In (c, d, g, h), blue dots and orange dots are for the weight feature distribution at the initial and final training stages, respectively. The top row is the result of two-layer networks, with the dropout layer after the hidden layer. The bottom row is the result of three-layer networks, with the dropout layer between the two hidden layers and after the last hidden layer. From Zhang and Xu \cite{zhang2024implicit}. \label{fig:condense_tanh}}
\end{figure}

\begin{figure}[h]
	\centering
	\subfloat[loss]{\includegraphics[height=0.32\textwidth]{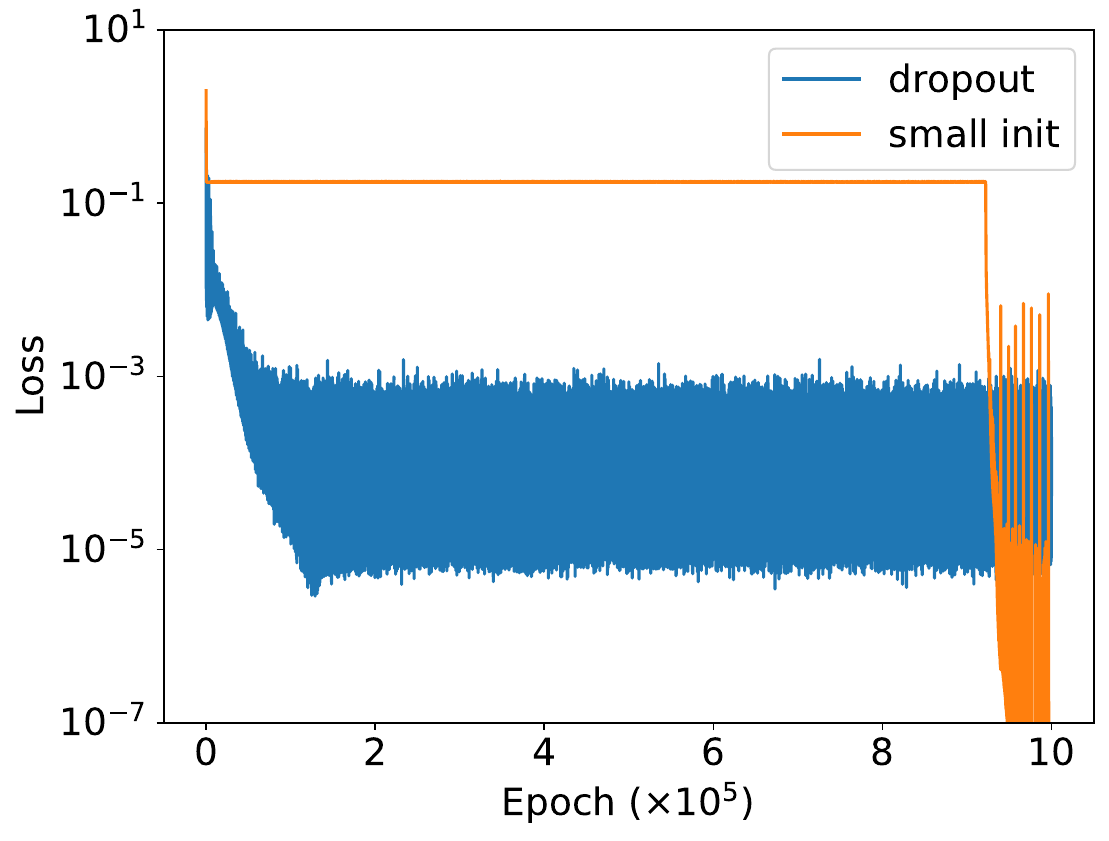}}
	\subfloat[output]{\includegraphics[height=0.32\textwidth]{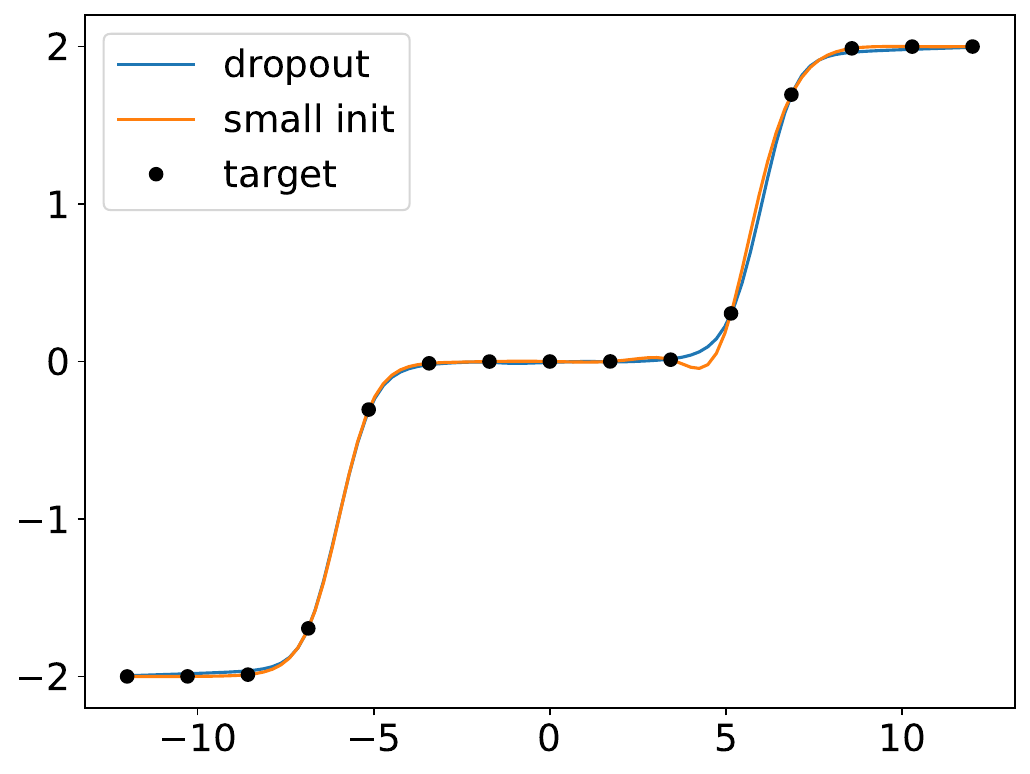}}

  \caption{Comparison of loss and output between the model trained by gradient descent with small initialization (orange) and the model trained by dropout with normal scale initialization (blue). The setup is the same as Fig. \ref{fig:condense_tanh}. From Zhang and Xu \cite{zhang2024implicit}.\label{fig:model_compare}}
\end{figure}

Previous sections demonstrate that neural networks exhibit condensation during training when employing small initialization. However, experiments in Fig. \ref{fig:energy_bar} suggest that this initialization approach, contrary to standard practices, may significantly slow network convergence and increase computational training costs. \cite{zhang2024implicit} reveals a compelling alternative: implementing dropout naturally induces network condensation, even without small initialization, as illustrated in Fig.~\ref{fig:condense_tanh}. Moreover, as demonstrated in Fig.~\ref{fig:model_compare}, dropout not only facilitates network condensation but also enables more rapid convergence to the ideal loss compared to small initialization. This approach significantly accelerates the model's learning dynamics while maintaining the desired condensation characteristics.


An intuitive explanation for dropout-induced condensation stems from its stochastic neuron elimination mechanism. During training, a subset of neurons is randomly deactivated, with the remaining neurons compensating for the eliminated ones. Upon convergence to an ideal solution, the surviving neurons at each step should play similar functions to the eliminated one in order to maintain functionally equivalent representations. Ideally, this process results in neurons with similar output functions.



\section{Subsequent works on condensation}
\subsection{Optimistic estimate}
In traditional learning theory, one often constrains model complexity to enhance generalization ability \cite{bartlett2002rademacher}. However, the classical theoretical approaches provide only loose generalization error bounds for NNs, primarily due to their over-parameterization with respect to the samples, resulting in a substantial discrepancy between theoretical predictions and practical training outcomes. Moreover, our observations of network condensation during training reveal that the effective parameters of neural networks are much fewer than their superficial parameters. Estimating the samples required for neural networks to achieve good generalization is an important problem.

\cite{zhang2023optimistic} introduces a method called \textbf{optimistic estimate} for estimating the required sample size in neural networks. The research reveals that the number of samples capable of recovering the target function is fundamentally linked to the intrinsic minimum width necessary for a neural network to represent that function. Moreover, this kind of generalization can be realized through network condensation. This demonstration suggests that expanding the width of neural networks does not increase the required number of samples and maintains their generalization ability.

\subsection{Reasoning ability of Transformer}
\cite{zhang2024initialization,zhang2025complexity} explore the role of condensation in enhancing the reasoning ability of Transformer models. The task is to study a composite function composed of several simple functions, i.e., addition and subtraction. Specifically, we define 4 simple functions (denoted as function 1, 2, 3, 4) and they can form 16 composite functions. We use 14 composite functions for training and leave the composition of functions 3 and 4 for testing (i.e., (3, 4) and (4, 3)). In distribution (ID) generalization refers to the accuracy of training composite functions with unseen computed numbers, while out of distribution (OOD) refers to the accuracy of test composite functions.

\begin{figure}
    \centering
    \includegraphics[width=0.95\linewidth]{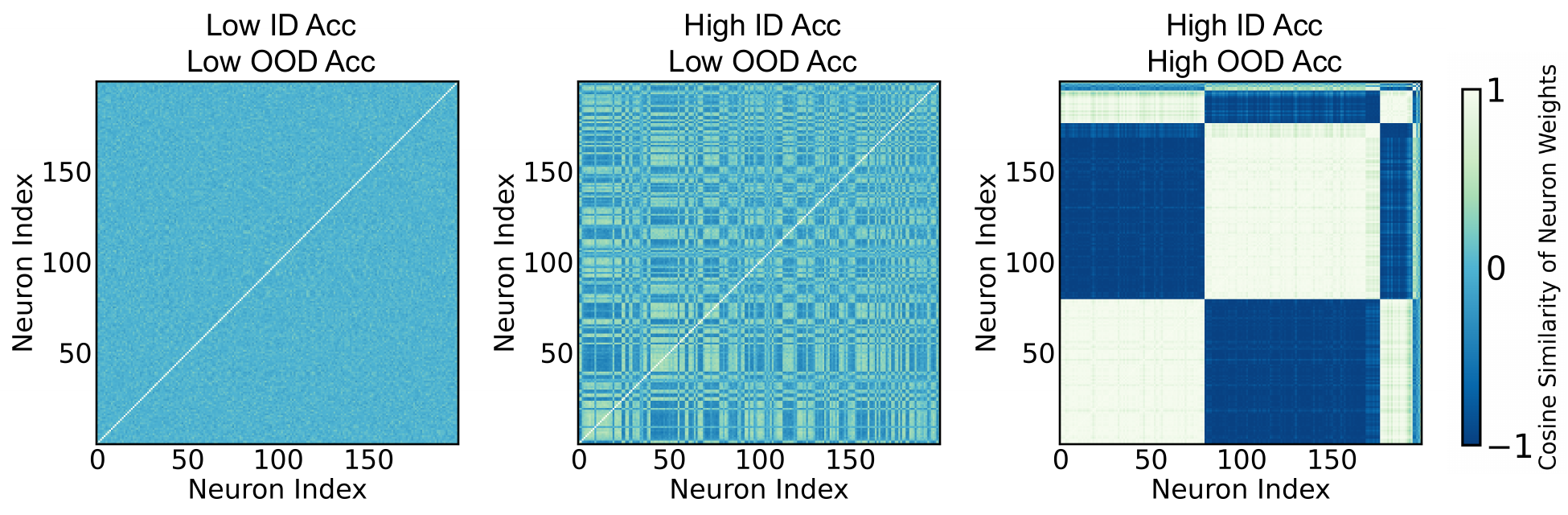}
    \caption{Cosine similarity matrices of neuron input weights (\(W^{Q(1)}\)). The abscissa and ordinate both represent the neuron index. The matrices are computed under the settings where the weight decay coefficient is fixed at 0.01, and the initialization rate (\(\gamma\)) is set to 0.2, 0.5, and 0.8 from the left panel to the right panel. }
    \label{fig:reasoning}
\end{figure}

The parameters of the transformer network are initialized by $\mathcal{N}(0,(\frac{1}{m^{\gamma}})^2)$, where $\frac{1}{m^{\gamma}}$ is the standard deviation and $m$ is the width of the layer. We observe that as the initialization rate $\gamma$ increases, i.e., initialization scale decreases, the transformer network learns the data respectively by the following four patterns: i) The network only remembers training data, and shows no generalization on any test data of seen or unseen composite functions; ii) The network can generalize to the seen composite function operating on unseen numbers, but not on the solution of unseen composite function (3, 4) or (4, 3), in addition, the network output of composite function (3, 4) and (4, 3) shows no symmetry; iii) Similar to (ii) but the network output of composite function (3, 4) and (4, 3) is symmetric; iv) The network generalizes to all composite functions, which indicates the network learns all primitive functions. This simple experiment shows that $\gamma$ can well tune the network to bias towards memorizing or reasoning data. 
Additionally, as shown in Fig.~\ref{fig:reasoning}, we notice that during this process, the phenomenon of condensation becomes increasingly pronounced, suggesting a strong correlation between the condensation phenomenon and the model's reasoning ability. A straightforward rationale is as follows: since the network strongly favors condensation, it tends to learn the data with the lowest possible complexity. Clearly, if the model can uncover the underlying simple function, it only needs to memorize a few simple functions rather than numerous data pairs. Consequently, it can explain the data with minimal effective complexity. An analysis of the initial training stage for reasoning bias of language models with small initialization further enhances the relation between condensation and reasoning \cite{yao2025analysis}. 


\subsection{Reduction of network width}
An approach to reduce a trained network can be readily proposed \cite{zhang2021embedding}. If a neural network is in an extremely condensed state, neurons within the same layer that share the same output function can be replaced by a single equivalent neuron. This equivalent neuron would have the input weights of the original neurons and an output weight that is the sum of the output weights of the original neurons. Consequently, the original neural network can be reduced to a much narrower network, thereby saving computational costs during the inference stage. \cite{chen2024efficient} utilize this reduction method for learning combustion problems, employing neural networks to solve ODEs through a data-driven approach. However, it should be noted that if a neural network is not in an extremely condensed state, such reduction can potentially harm performance, depending on the degree of condensation. Continuous training of the reduced network can mitigate this harm.

\section{Discussion}
The condensation phenomenon has been observed during the training of simple two-layer neural networks and has since been extended to more complex architectures, such as convolutional neural networks and Transformer networks. While condensation is a common feature during nonlinear training, it should not be expected to manifest as an extremely condensed state in every case. Condensation is rather a tendency or bias during nonlinear training that can be enhanced or suppressed depending on the choice of hyperparameters and optimization tricks. Condensation represents a distinctive viewpoint on DNNs, intimately connected to the model architecture. This perspective introduces features that surpass those found in traditional machine learning techniques, including kernel methods, and contrasts with other views like low-frequency bias and the flatness/sharpness of minima. 

The condensation phenomenon provides valuable insights into the behavior of neural networks, from their generalization capabilities to their reasoning abilities. However, the study of condensation is still in its early stages. In the future, we anticipate significant theoretical advancements and practical approaches to harness the condensation effect for more effective utilization of neural networks.

\bibliographystyle{alpha}
\bibliography{ref}

\end{document}